\documentclass[num-refs]{wiley-article}

\usepackage{siunitx}
\usepackage{graphicx}
\usepackage{subcaption}
\usepackage{amsmath} 
\usepackage{amssymb}  
\usepackage{tikz}
\usepackage{multirow}
\usepackage{soul}
\usepackage{hyperref}
\papertype{Original Article}
\paperfield{Journal Section}

\title{ViVa-SAFELAND: a New Freeware for Safe Validation of Vision-based Navigation in Aerial Vehicles}

\abbrevs{EAV, Emulated Aerial Vehicle; UAV, Unmanned Aerial Vehicles; CNN, Convolutional Neural Network; SLZ, Safe Landing Zone; SS, Semantic Segmentation; VAC, Virtual Aerial Camera; FoV, Field Of View; RoI, Region of Interest; EDSR, Enhanced Deep Super-Resolution; YOLO, You Only Look Once; SORA, Specific Operations Risk Assessment; RL, Reinforcement Learning.}

\author[1]{Miguel S. Soriano-García}
\author[2]{Diego A. Mercado-Ravell}

\affil[1]{Center for Research in Mathematics CIMAT, campus Zacatecas,
        Calle Lasec y Andador Galileo Galilei, Manzana 3, Lote 7, 98160 Zacatecas, Mexico. miguel.garcia@cimat.mx}
\affil[2]{Centro de Investigación y de Estudios Avanzados del IPN, CINVESTAV-IPN unidad Guadalajara, Av. del Bosque 1145, 45019, Zapopan, Jalisco, Mexico. diego.mercado@cinvestav.mx}

\corraddress{Diego A. Mercado-Ravell}
\corremail{diego.mercado@cinvestav.mx}

\fundinginfo{This work was supported by Office of Naval Research Global ONRG, Award No. N62909-24-1-2001.}

\runningauthor{Soriano-Garcia et al.}

\begin{document}

\begin{frontmatter}
\maketitle

\begin{abstract}
ViVa-SAFELAND is an open source software library, aimed to test and evaluate vision-based navigation strategies for aerial vehicles, with special interest in autonomous landing, while complying with legal regulations and people's safety. It consists of a collection of high definition aerial videos, focusing on real unstructured urban scenarios, recording moving obstacles of interest, such as cars and people. Then, an Emulated Aerial Vehicle (EAV) with a virtual moving camera is implemented in order to ``navigate" inside the video, according to high-order commands. ViVa-SAFELAND provides a new, safe, simple and fair comparison baseline to evaluate and compare different visual navigation solutions under the same conditions, and to randomize variables along several trials. It also facilitates the development of autonomous landing and navigation strategies, as well as the generation of image datasets for different training tasks. Moreover, it is useful for training either human of autonomous pilots using deep learning. The effectiveness of the framework for validating vision algorithms is demonstrated through two case studies, detection of moving objects and risk assessment segmentation. To our knowledge, this is the first safe validation framework of its kind, to test and compare visual navigation solution for aerial vehicles, which is a crucial aspect for urban deployment in complex real scenarios.

\end{abstract}
\end{frontmatter}

\section{INTRODUCTION}

The use of Unmanned Aerial Vehicles (UAVs) has rapidly grown in various civilian areas, such as precision agriculture, package delivery and recreational activities; thanks to their high mobility, maneuverability, and low manufacturing cost \cite{Shakhatreh}. However, despite huge technological advances, their potential is limited in urban areas  due to the risk of accidents that can harm people. This has restricted their use mainly to controlled or rural environments far away from people, which is also reflected in very restrictive legal regulations \cite{SORA2019}. In densely populated areas, the risk of UAV failures, such as human errors, loss of communication or power shortage, remains a serious concern, where a drone crash may result in severe injuries or fatal accidents, not to mention material damage, in spite of current safety policies like return-to-home protocols. To enable safer use of these devices in urban environments, it is important to develop autonomous landing algorithms able to reduce the risk of accidents in case of system failure. This is essential to reduce injury risks and guarantee public safety, and will help to unleash the true potential of UAVs in urban environments \cite{Haoran}.

Accordingly, with the great recent success of deep learning solutions for visual perception, vision-based autonomous landing has gained increasing attention, specially in complex urban areas. A possible solution is the use of fiducial markers to identify safe landing zones like in the works at \cite{lv2024autonomous, wang2022quadrotor}. In \cite{Pieczyński}, authors further employ lightweight neural networks for human presence detection around the landing pad. Nevertheless, the availability of such previously marked zones in emergency landing situations is not guaranteed. Hence, autonomous landing in urban unstructured scenarios is a key challenge, as established  by several studies \cite{Kakaletsis2022, SHAHALAM2021115091} that emphasize the importance of computer vision and the integration of various methods for detecting safe landing zones. These approaches enhance the resiliency and robustness of autonomous flight operations, ensuring greater safety in complex environments such as urban areas. 

One approach is to detect planar regions on the image, for instance the study in \cite{KALJAHI2019319} uses Gabor Transform and Markov Chain Codes to identify planar regions as candidate landing zones, but only validation upon image datasets was provided. Also, people detection and tracking is a critical aspect to prevent accidents, in \cite{gonzalez2021lightweight}, a lightweight Convolutional Neural Network (CNN) was proposed in order to infer the people's position in densely populated scenarios. The proposed algorithm was tested in real-time on a low cost processor with a human operated UAV. Similarly, \cite{Tzelepi2021} uses lightweight deep neural networks to detect human crowds in drone-captured images. It generates heat maps and applies a graph embedding-based regularization technique. Afterwards, in \cite{GonzalezRAL_s} the authors developed a Safe Landing Zone (SLZ) detection algorithm, where multiple candidate SLZ free of people where detected and tracked along time. Nevertheless, only accidents directly involving people where taken into account, and these algorithms where only tested out of line in pre-recorded videos and in image datasets. Meanwhile, \cite{Mitroudas2023} utilizes object detectors and OctoMap to generate point clouds, identify obstacle-free areas, and determine safe landing spots, whereas the results were validated in the Gazebo simulator.
\newpage

In order to capture more information from the scene context, rather than only searching people, Semantic Segmentation (SS) algorithms had been proposed. Using SS techniques, a CNN can be trained to infer at pixel level the class and localization of different objects of interest, in the context of autonomous landing, the more common labels being people, cars, grass, vegetation, animals, pavement, dirt, etc. In \cite{ahmed2021real} a real-time SS system for aerial images based on the U-Net architecture, combined with MobileNet, achieved good accuracy in the segmentation task, standing out for its effectiveness and speed. Another relevant approach is the Self-Attention based Graph Neural Network SAGNN, which outperformed benchmark methods in the SS of aerial images, demonstrating its ability to improve object detection accuracy from drones \cite{diao2022superpixel}. Similarly, \cite{Benjwal} enhances low-resolution images with a super-resolution model before performing image segmentation, achieving significant improvements in accuracy. In \cite{Abdollahzadeh}, SS is applied to create continuous safety maps, showing promising results on public datasets. \cite{Zhang24} employs an improved U-Net architecture to evaluate safe zones at different descent altitudes, demonstrating effectiveness through validation with public datasets. Moreover, \cite{chuyalex} employs visual transformers for SS and risk assessment in the context of autonomous landing in unstructured urban scenarios. Unfortunately, these works are constrained to the evaluation of the CNN architectures for the SS task, but do not attempt to solve or evaluate the autonomous landing nor the visual navigation problems. Alike, the work at \cite{Mitroudas} incorporates a multi-criteria decision-making module to assess the risk posed by obstacles during emergency drone landings. The system was tested using the Gazebo simulator in scenarios involving both stationary and moving individuals, demonstrating robustness across various test conditions. In \cite{vemulapalli2024reinforcement} a study applied Reinforcement Learning (RL) to optimize the autonomous landing of quadcopters in a simulated environment using AirSim and Unreal Engine.


As observed, developing vision based autonomous landing strategies in unstructured environments is a very recent and challenging task, where a lot of work is still required, provided that the scenes are in general extremely diverse, complex, unstructured and unknown, and there may be plenty of moving obstacles that pose great risk of accidents, such as people and cars. Moreover, testing and validating these algorithms in real populated areas, where they are expected to function eventually, can be as challenging and dangerous, or even more, than the problem they aim to solve. Therefore, it is crucial for these algorithms to be validated in controlled environments or through virtual simulations. So far the main validation approach is the use of virtual reality engines for simulation, mainly Gazebo or AirSim coupled with Unreal engine. Another validation tool in this context, is the work at \cite{tovanche2024VR} which employs a "robot-in-the-loop" approach combined with AirSim and Unreal. In this method, a virtual avatar mimics the movements of a real drone, but performs visual perception and navigation on a virtual urban scenario with moving virtual agents. An important limitation of these approaches is the virtual-to-reality gap, since the visual perception and decision-making occurs in a virtual environment, which can lead to significant differences when transferring these algorithms to real-world situations where environments are more complex, diverse and unpredictable. To our knowledge, no tool currently exists to evaluate or compare different vision navigation algorithms for UAVs, under equal conditions, without using virtual simulators.

With these in mind, it is clear the lack of validation tools that allow to test vision-based navigation techniques over real-world scenarios. In response, this work presents the Vision Validation Safe Landing (ViVa-SAFELAND) framework, an open source software tool designed for safe validation of vision navigation solutions in unstructured real environments, by dynamically simulating a UAV over previously recorded videos of various real urban scenes. These videos are recorded with a drone at a high safe altitude, from a static position, while capturing moving obstacles in the scene, including people, vehicles, etc. Then, an Emulated Aerial Vehicle (EAV) equipped with a Virtual Aerial Camera (VAC) is implemented to navigate within the video, while capturing real snapshots within the larger video. The EAV is controlled by high level commands (desired attitude and total thrust), given either by a human user or by an autonomous navigation algorithm. As a proof of concept, two case studies are presented to demonstrate the potential of the framework for evaluating vision algorithms. On the one hand, a risk assessment algorithms is implemented using SS with a U-NET architecture. On the other hand, a YOLO (You Only Look Once) based network is also implemented and tested to detect in real time critical moving obstacles such as vehicles and people, effectively showcasing the potential of the framework to evaluate different vision techniques.  

This approach allows for the safe and efficient validation and development of vision-based algorithms for UAV missions in civilian applications, including autonomous and emergency landing strategies. By not requiring direct testing with UAVs in populated environments, the risks of accidents and third-party damage are significantly reduced, also complying with legal regulations governing the use of these vehicles in urban areas. Moreover, the implementation of the UAV's equations of motions, makes it a suitable tool for human pilot and deep learning based autonomous pilot training, enhancing their visual navigation skills in conditions that faithfully capture real-world scenarios.

The main contributions of this work are summarized as follows: \begin{itemize}
    \item ViVa-SAFELAND provides a simple, easy to use and fair baseline for comparison among different visual based navigation algorithms, for example, for autonomous landing in complex urban scenarios.
    \item It is a powerful tool for safely developing and testing vision based navigation algorithms in real environments, where real UAVs could not operate due to legal restrictions, such as in populated or urban areas.
    \item It allows for training human and automatic pilots by controlling an emulated aerial vehicle with a virtual camera, that faithfully reproduces movements over real scenes.
    \item It will help with the automatic generation of labeled datasets for different training tasks.
\end{itemize}

The reminding of the paper is organized as follows, Section \ref{sec:framework} describes the validation framework, explaining how the UAV is emulated considering its equations of motion, as well as the virtual aerial camera. Section \ref{sec:study} presents two case studies testing in real-time a SS and an object detection algorithms on the images obtained by the EAV in motion. Finally, Section \ref{sec:conclusions} offers conclusions and future research directions.

\section{ViVa-SAFELAND VALIDATION TOOL}
\label{sec:framework}

\begin{figure}
    \centering
    \includegraphics[width=\linewidth]{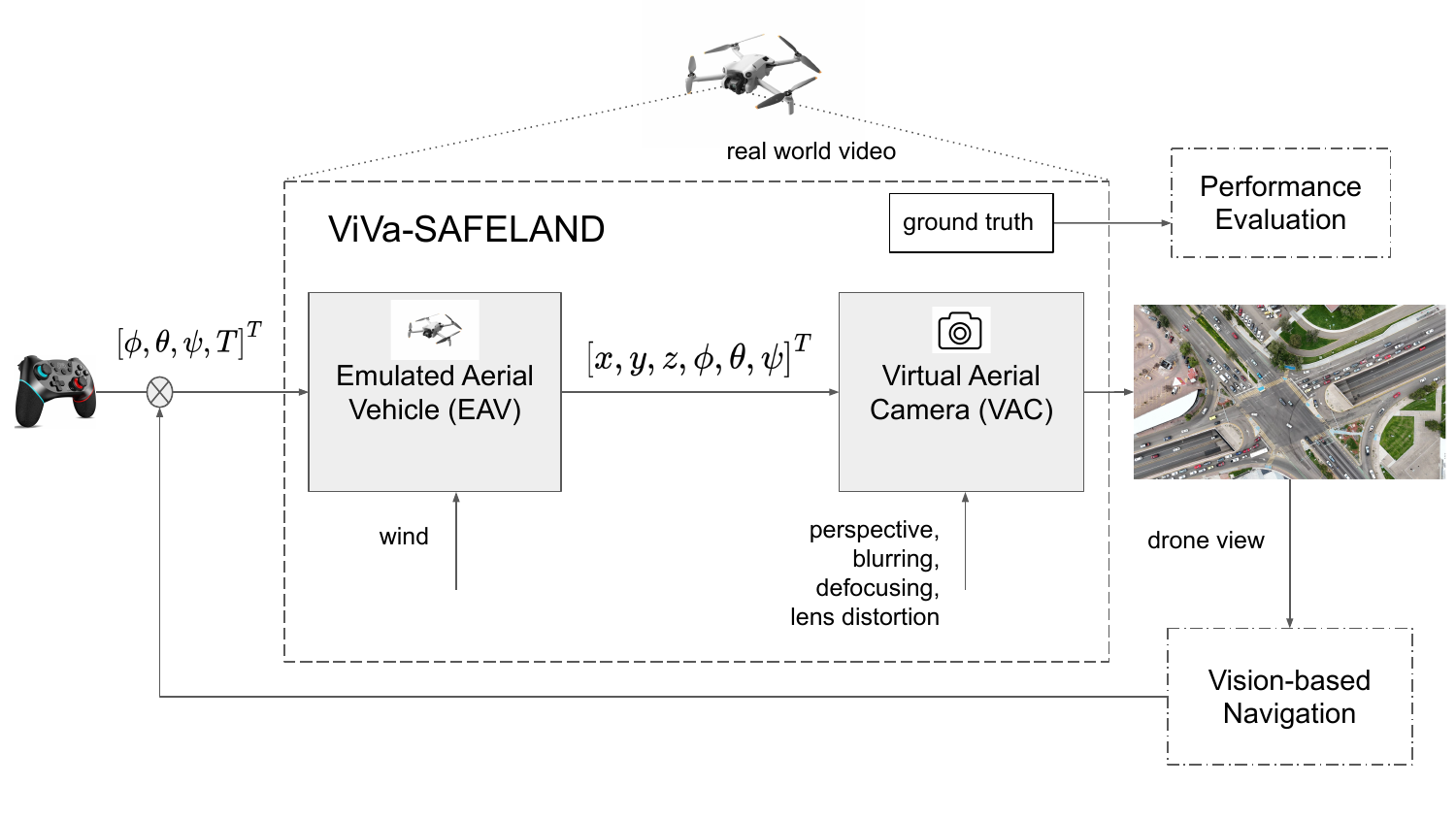}
    \caption{ViVa-SAFELAND: a Visual Validation Safe Landing tool. An Emulated Aerial Vehicle is implemented to navigate inside aerial videos of real urban environments, from where the drone's view is recovered and used to evaluate vision algorithms.}
    \label{fig:blocks}
\end{figure}

This work presents the Visual Validation Safe Landing framework ViVa-SAFELAND (see Fig. \ref{fig:blocks}), a platform developed in \textit{Python 3.10}, designed for validation, training, and safe testing of vision algorithms related to aerial vehicles. It uses the \textit{pygame 2.5.2} library for graphical simulation and user interaction, \textit{numpy 1.26.4} for efficient numerical calculations, and \textit{Opencv 4.10.0.84} for image processing and UAV camera modeling. This framework allows simulating the dynamic behavior of a UAV over previously recorded real scenarios, facilitating the control of the EAV through a keyboard, joystick, or an autonomous navigation algorithm, providing an authentic UAV handling experience.

The reference videos used in this work were captured using a \textit{DJI Mini 4 Pro} drone, weighting 0.25Kg, equipped with a 48 MP camera with a gimbal, which was oriented downwards ensuring that the ground plane is mainly parallel to the image plane. This drone is capable of hover flight and video stabilization, resulting in a very stable video from a quasi static position. The videos were recorded in 8K resolution at 30 fps with a Field Of View (FoV) angle of $82.1^o$. The recordings were made in various urban locations across Mexico. We provide around 20 videos freely available, including diverse urban environments, with vehicular and pedestrian movement. The relative flight altitudes during the recordings ranged from 50m to 220m.

Its main purpose is to enable the validation of vision navigation and autonomous landing algorithms, as well as human pilots training, automatic labeled datasets generation and training of deep learning navigation solutions, in a safe environment without accident risks. Thanks to its ability to conduct tests in scenarios prohibited by real-world regulations, such as urban areas, this framework opens up the possibility of safely and effectively developing and improving deep learning models and other algorithms, thereby increasing the potential for UAV's civilian applications.

\subsection{Emulated Aerial Vehicle (EAV)}

Consider an Emulated Aerial Vehicle (EAV) living inside the pre-recorded video stream, with mass $m$, and center of mass with coordinates $\mathbf{X_k}= [x_k,y_k,z_k]^T \in  \mathbb{R}^3$ at time $k$, with respect to an inertial reference frame. Then, the dynamics of motion of the EAV can be modeled by the second order discrete equation of the form \cite{mahony2012multirotor}

\begin{equation}
        \mathbf{X_k} \approx  2 \mathbf{X_{k-1}} - \mathbf{X_{k-2}} +   \frac{\Delta_t^2}{m} \left( \mathbf{R_k} T_k \mathbf{e_3} - mg\mathbf{e_3} - \frac{k_d}{\Delta_{t}} \left( \mathbf{X_{k-1}} - \mathbf{X_{k-2}} \right) + \boldsymbol{\xi_k} \right) 
\end{equation}

where: 
\begin{itemize}
    \item \(\mathbf{X_k}=[x_k,y_k,z_k] \in \mathbb{R}^3\): drone's position at time $k$.
    \item \( \Delta_t \in \mathbb{R}\): time period.
    \item \( m \in \mathbb{R}\): mass of the vehicle.
    \item \( \mathbf{R_k} \in SO(3)\): rotation matrix from the body frame to the inertial reference frame.
    \item \( T_k \in \mathbb{R}\): total thrust force.
    \item \( \mathbf{e_3}=[0, 0, 1]^T \in \mathbb{R}^3\): unit vector.
    \item \( g \in \mathbb{R}\): gravity constant.
    \item \( k_d \in \mathbb{R}\): air drag coefficient.
    \item \( \boldsymbol{\xi_k} \in \mathbb{R}^3\): external disturbances.
\end{itemize}

The translational movement of the EAV is driven by the total thrust force $T_k$ and the rotation matrix $\mathbf{R_k}(\phi,\theta,\psi)\in SO(3)$, which is parameterized by the Euler angles roll $\phi$, pitch $\theta$ and yaw $\psi$, which serve as the control inputs for the EAV. Note that this model and operation mode is in compliance with a UAV with a well tuned internal attitude controller, which is common in most commercial drones nowadays.

\subsection{Virtual Aerial Camera (VAC)}
Let us define the original video from the real world, as a sequence if images $\mathcal{I}^o_k(u_k,v_k) \in \mathbb{R}^{W\times H}$ where $u_k,v_k$ are pixel coordinates. Then, knowing the original camera altitude $z_o$, the image width $W$ and height $H$, and intrinsic parameters, it is the focal length $f_o$, pixel width and height $ \rho_{wo},\rho_{ho}$ and principal point coordinates $u_{o0},v_{o0}$, considering also a central camera model \cite{corke17} and assuming that all the points in the image $\mathcal{I}^o_k$ belong to a plane parallel to the image, then, the set of points in the virtual world corresponding to the original video with origin at the center of the image are obtained as 
\begin{equation}
    \mathbf{\Tilde{p}_w}=\frac{z_o}{f_o}[u_k\rho_{wo}+\frac{W}{2},v_k\rho_{ho}+\frac{H}{2},0,\frac{f_o}{z_o}]^T\in \mathbb{R}^4
\end{equation}
Note that the real scale is easily recovered since the recording altitude $z_o$ is known.

Then, once simulating the EAV, we can incorporate a down looking Virtual Aerial Camera (VAC), rigidly attached to the EAV. The homogeneous matrix defining the rigid transformation between the virtual world and the VAC reference frame $\mathbf{H^w_v}\in SE(3)$ is given by:

\begin{equation}
    \mathbf{H^w_v}=\mathbf{H^v_w}^{-1}=
    \left[\begin{array}{cc}
         \mathbf{R_k}& \mathbf{X_k}\\ \left[0 \ 0\  0\right] & 1 
    \end{array}
    \right]
    \left[\begin{array}{cc}
         \mathbf{R^b_v}& \mathbf{T^b_v}\\ \left[0 \ 0\  0\right] & 1 
    \end{array}
    \right]
\end{equation}
where $\mathbf{R_k} \in SO(3)$, and $\mathbf{X_k}$ are again the pose (orientation and position) of the EAV, with respect to the virtual world frame. $ \mathbf{R^b_v}\in SO(3)$ and $\mathbf{T^b_v}\in \mathbb{R}^3$ correspond respectively to the rotation matrix and translation vector from the VAC frame to the EAV body frame.

Now we can compute the VAC image's width $w_k$ and height $h_k$ in pixels, by using the VAC's FoV, considering a horizontal and vertical angles of view $\{\gamma_h, \gamma_v \}\in \mathbb{S}^2$:

\begin{equation}
    \begin{array}{ccc}
        w_k = 2z_k\tan{(\frac{\gamma_w }{2}) } & ; &  h_k = 2z_k\tan{(\frac{\gamma_h }{2} )} 
    \end{array}
\end{equation}
then, the VAC image $\mathcal{I}^{vac}_{k}$ at time instant $k$ can be obtained as a Region of Interest (RoI) from the original video with a perspective transformation, and it is defined as the set of all points $\mathbf{\Tilde{p}_i}=[\mathrm{x}_i,\mathrm{y}_i,0,1]^T\in \mathbb{R}^4$ in the virtual world, projected to the VAC image plane, it is:

\begin{equation}
    \mathcal{I}^{vac}_{k}=\left\{ \mathbf{C_v} \mathbf{H^v_w}\mathbf{\Tilde{p}_i}\ \left|\begin{array}{l} \vert\mathbf{\Tilde{p}_i}(\mathrm{x}_i)-\mathbf{\Tilde{p}_E}(\mathrm{x}_k)\rvert \leq w_k  \wedge \  \vert\mathbf{\Tilde{p}_i}(\mathrm{y}_i)-\mathbf{\Tilde{p}_E}(\mathrm{y}_k)\rvert \leq h_k
    \end{array} \right.\right\}
\end{equation}
the intrinsic VAC matrix $\mathbf{C_v}(f, \rho_w, \rho_h, u_0, v_0)\in \mathbb{R}^{3\times4}$ is parameterized respectively by the focal length, the pixels width and height, and the principal point coordinates. Note that different VAC models can be implemented to acknowledge for different cameras and capture lens distortion effects. Moreover, $\mathbf{\Tilde{p}_E}=[\mathrm{x}_k,\mathrm{y}_k,\mathrm{z}_k,1]^T\in \mathbb{R}^4$ stands for the projection of the VAC optical axis to the horizontal plane of the virtual world, it is:

\begin{equation}
    \mathbf{\Tilde{p}_E}=\mathbf{H^w_v}
        [0 \ 0 \ \frac{-z_k}{\mathbf{z_v}\cdot \mathbf{z_w}} \ 1]^T
\end{equation}
by considering the virtual world vertical axis $\mathbf{z_w}$ and the VAC frame vertical axis $\mathbf{z_v}$. 

Note that the VAC image $\mathcal{I}^{vac}_{k}$ may also be obtained using only image transformations over a RoI of the original video frames $\mathcal{I}^{o}_{k}$, using zoom, translations and warp perspective image transformations. Also, the framework can be extended to acknowledge other physic phenomena such as blurriness, lens distortion, vibrations, illumination changes, etc.

\subsection{Super-Resolution}
As the EAV approaches the ground, the VAC images tend to lose resolution and appear low-quality. To mitigate this issue, a super-resolution network based on \cite{lim2017enhanced} is employed to enhance the visual quality of the images captured by the VAC. This super-resolution approach is applied exclusively to the images obtained by the VAC, ensuring that they maintain an adequate level of detail even when the EAV is at low altitude.

\begin{figure}
    \centering
    \includegraphics[width=0.5\linewidth]{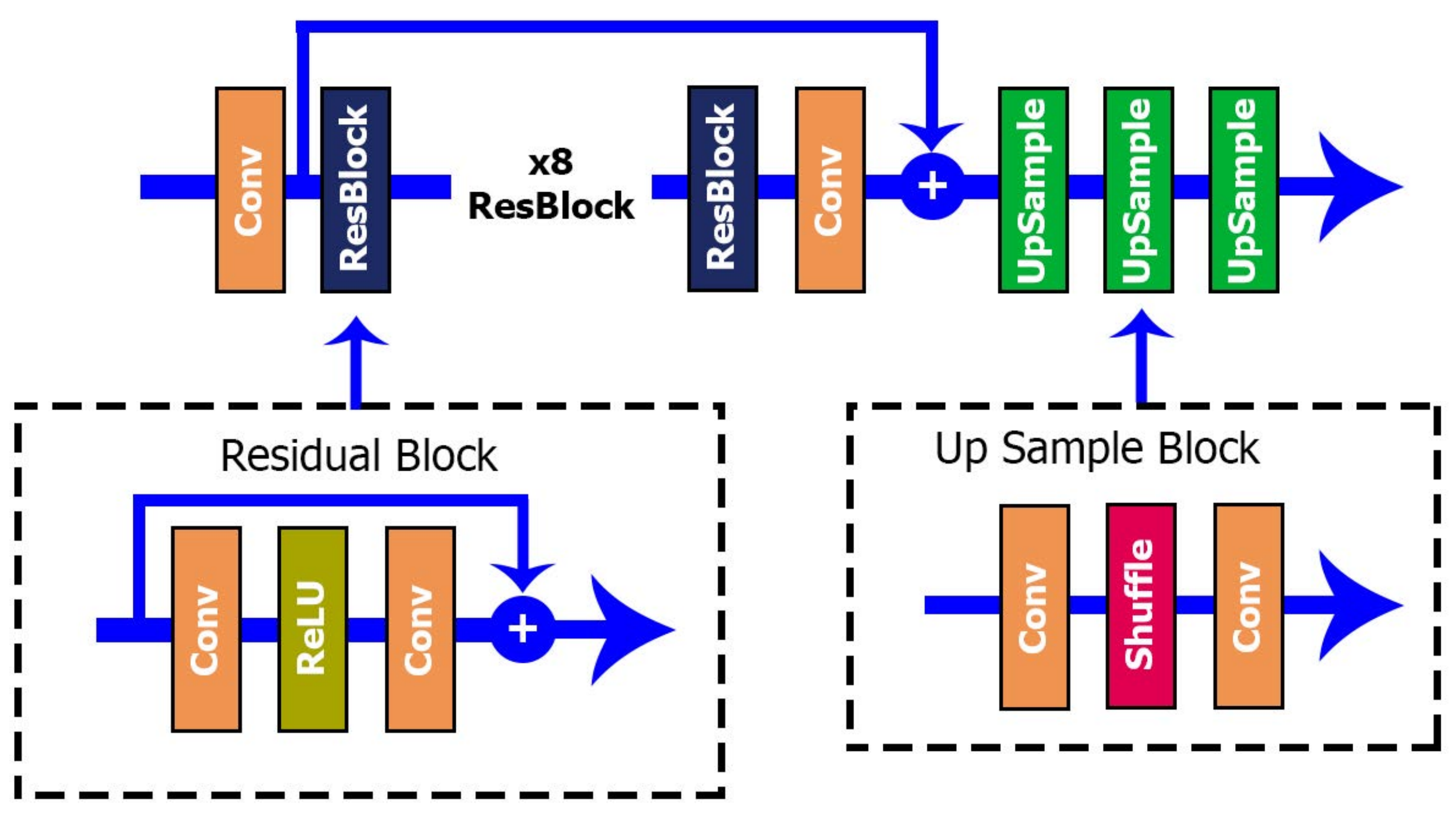}
    \caption{Architecture of the network used to apply super-resolution to the images captured by the VAC at low altitudes.}
    \label{fig:model}
\end{figure}

\begin{figure}
    \centering
    \begin{subfigure}{0.9\textwidth}
        \centering
        \includegraphics[width=.30\linewidth]{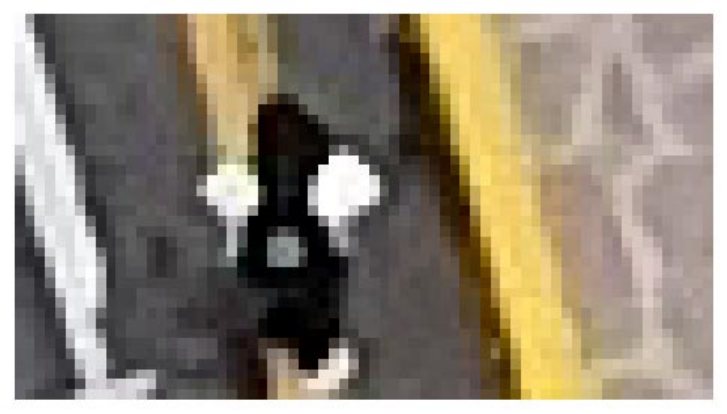}
        \includegraphics[width=.30\linewidth]{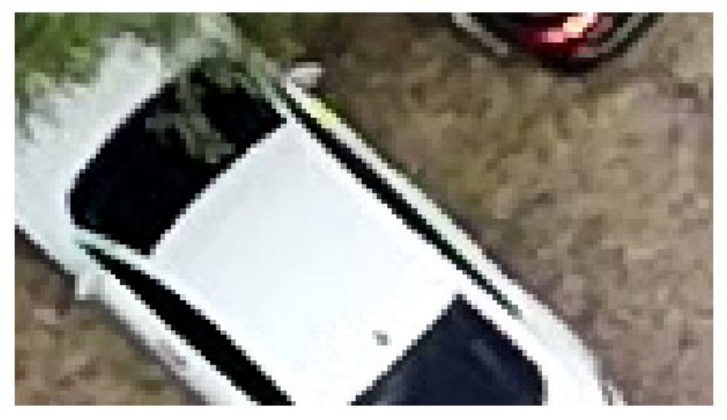}
        \includegraphics[width=.30\linewidth]{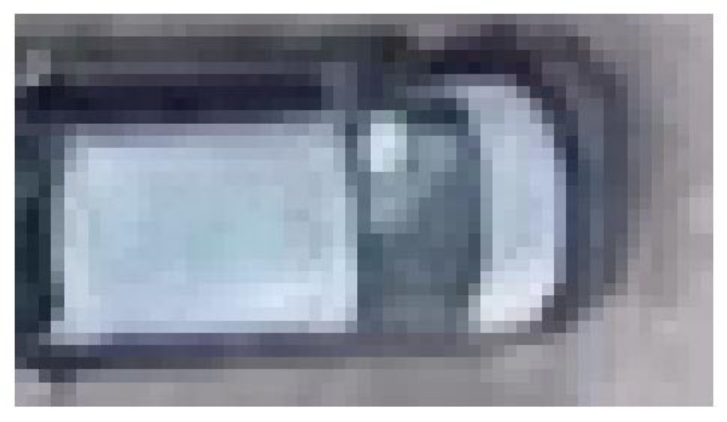}
    \end{subfigure}
    \begin{subfigure}{0.9\textwidth}
     \centering
        \includegraphics[width=.30\linewidth]{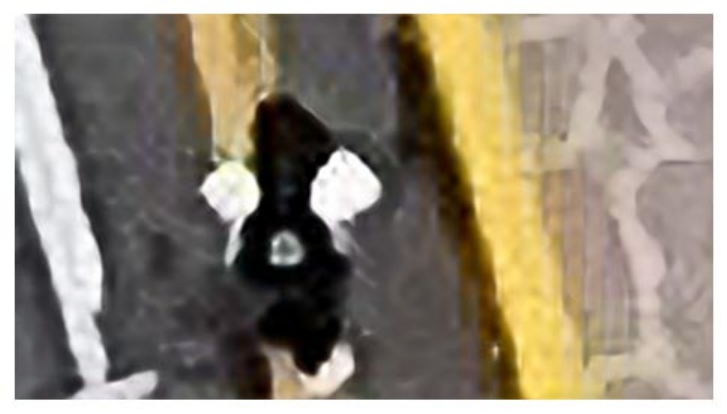}
        \includegraphics[width=.30\linewidth]{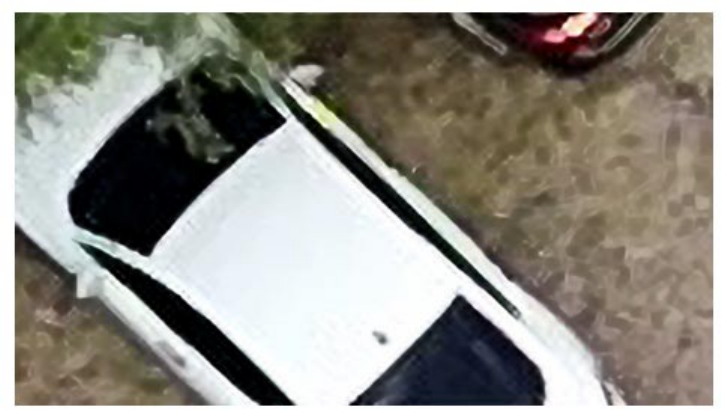}
        \includegraphics[width=.30\linewidth]{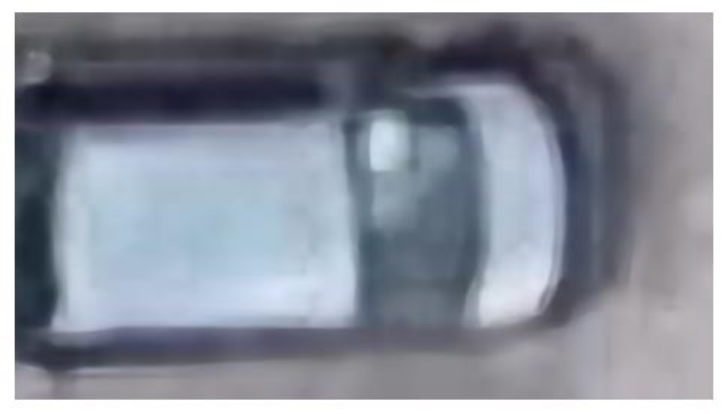}
    \end{subfigure}
    \caption{Comparison of the original images captured by the VAC (top) and the images restored by the super-resolution network (bottom) under different altitudes from 110m to 200m.}
    \label{fig:res}
\end{figure}

The architecture used is based on the Enhanced Deep Super-Resolution Networks (EDSR) model, whose structure is shown in Fig. \ref{fig:model}. The process begins with a convolution layer that transforms the input image, composed of 3 color channels, into a set of 64 features. Next, 8 residual blocks are used to process these features. At the end of the network, upsample blocks are included, each of which increases the image size by a factor of 2. To double the image size, a single upsample block is required. For larger increases, such as a factor of 8, three consecutive upsample blocks must be used, allowing a gradual increase in resolution.
In Fig. \ref{fig:res}, we show the results of the super-resolution model applied to some image samples captured by the VAC. On the top row, the original zoomed-in images seen by the VAC are displayed, while on the bottom row the images restored by the network are depicted. In the first column, the reference shot was taken at an altitude of 110m, while the VAC is 2m above the ground. In the center column, the original reference shot was taken at 110m, but the VAC is 4m above the ground. These subfigures show that the original images appear pixelated, but the model can restore them by detecting key features and improving visual details. However, in the right column, the reference shot was taken at 200m, while the VAC is 3m above the ground, resulting in a greater loss of detail in the original image. Despite this, the super-resolution model manages to recover some details and improves visual quality by eliminating pixelated areas, though with slightly lower accuracy compared to the previous cases. These improvements in the image quality may be key for the correct performance of computer vision and deep learning algorithms, henceforth, for the correct evaluation of vision based navigation strategies.

\section{EXPERIMENTAL RESULTS: OBJECT DETECTION AND RISK ASSESSMENT}
\label{sec:study}

This platform is designed to address the challenging problem of vision based navigation, with a particular interest in autonomous landing for UAVs, specially in complex real-world scenarios. Autonomous landing is inherently difficult due to various unpredictable factors such as wind, blurriness, complex and very diverse scenarios, and critical obstacles with unknown dynamics like vehicles, people, or animals. The platform allows for the safe validation, training, and testing of navigation and landing algorithms, as well as pilot training, in a controlled environment that effectively simulates these challenges. By capturing the movement of the UAV and the visual interaction with complex scenarios including moving obstacles, which is critical for validating autonomous navigation algorithms, it replicates the real-world conditions much better than a virtual simulator while preserving the system safety during the development and testing phase. Furthermore, it enables testing in scenarios where the use of real UAVs would be restricted due to safety regulations, such as urban environments. This framework facilitates the development and refinement of advanced algorithms, including those based on deep learning, in a secure and efficient manner, significantly enhancing the potential of UAVs
for real-world civilian applications.

\begin{figure}
    \centering
    \includegraphics[width=1\linewidth]{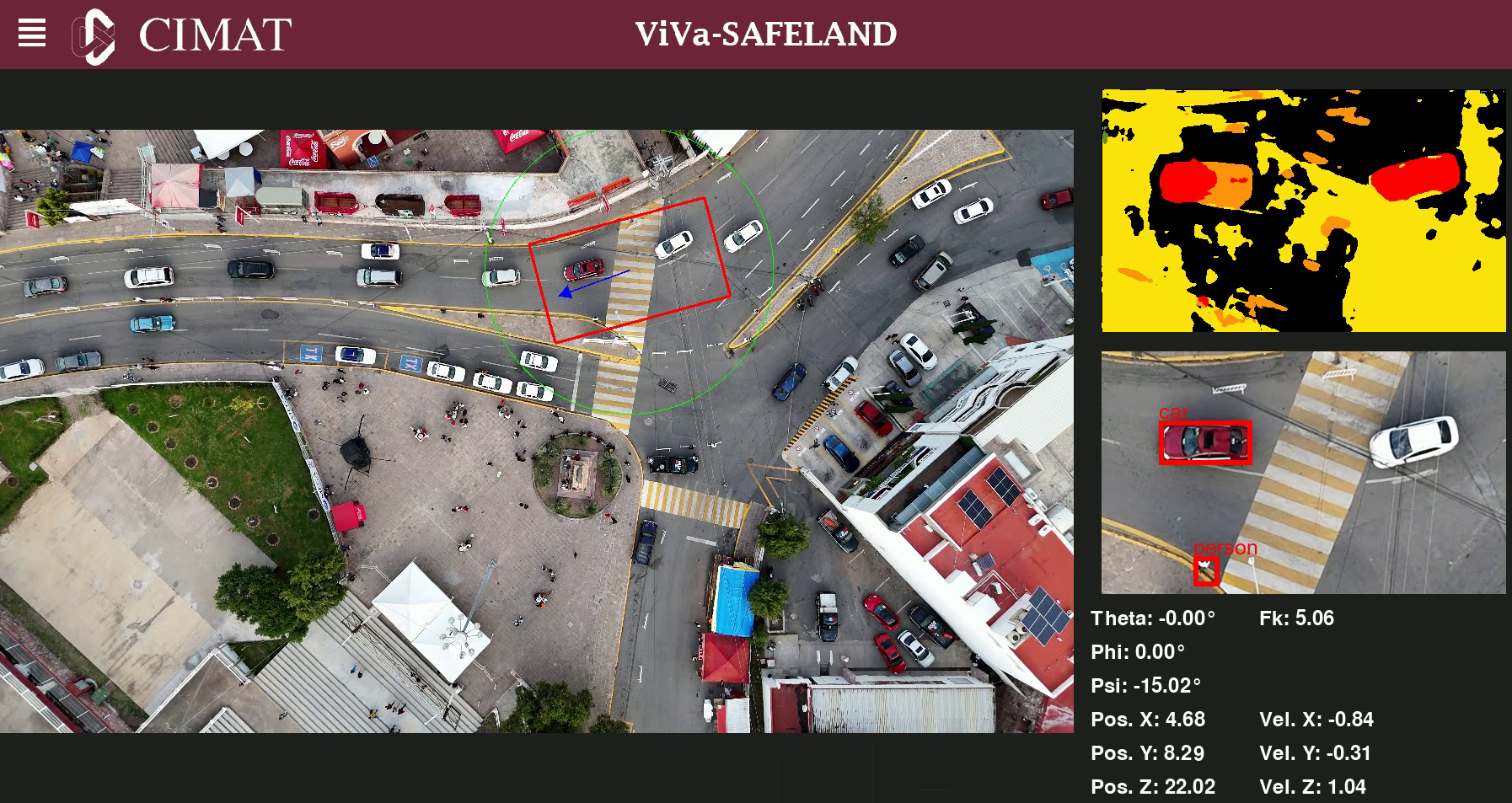}
    \caption{Example of ViVa-SAFELAND operation. The larger left image presents the original video frame, where the VAC RoI is represented by a red box, the EAV velocity vector is represented by the blue arrow for the horizontal coordinates, and as a green circle for the z-axis velocity. The VAC view is depicted in the central right box, where YOLO is implemented to identify critical obstacles such as people and cars. In the upper right box, a U-Net is used for risk assessment using image segmentation. At the bottom right, the drone's flight information is displayed.}
    \label{fig:viva}
\end{figure}

As a proof of concept, two case studies are presented to demonstrate the application of the framework, as depicted in Fig. \ref{fig:viva}. In the first case, a semantic segmentation U-Net network was trained to process the VAC view and assign at pixel level different risk levels. The second case study involved integrating a YOLOv8 (You Only Look Once) model to detect critical objects such as vehicles, people, motorcycles, and bicycles. This approaches allows for simultaneously evaluating the UAV's surroundings in terms of both object identification and risk assessment, which are crucial for improving decision-making in autonomous flight and landing missions. A video showcasing the ViVa-SAFELAND capabilities, and evaluating the two case studies can be watched at
\url{https://drive.google.com/file/d/1hkpAu1rIzMpKmuwgYh6kKyL0hU0SoiwG/view?usp=sharing}.

\subsection{Experimental Validation of Perception Algorithms}
\begin{figure}[h!]
    \centering
    \begin{subfigure}{\textwidth}
        \centering
        \includegraphics[width=.48\linewidth]{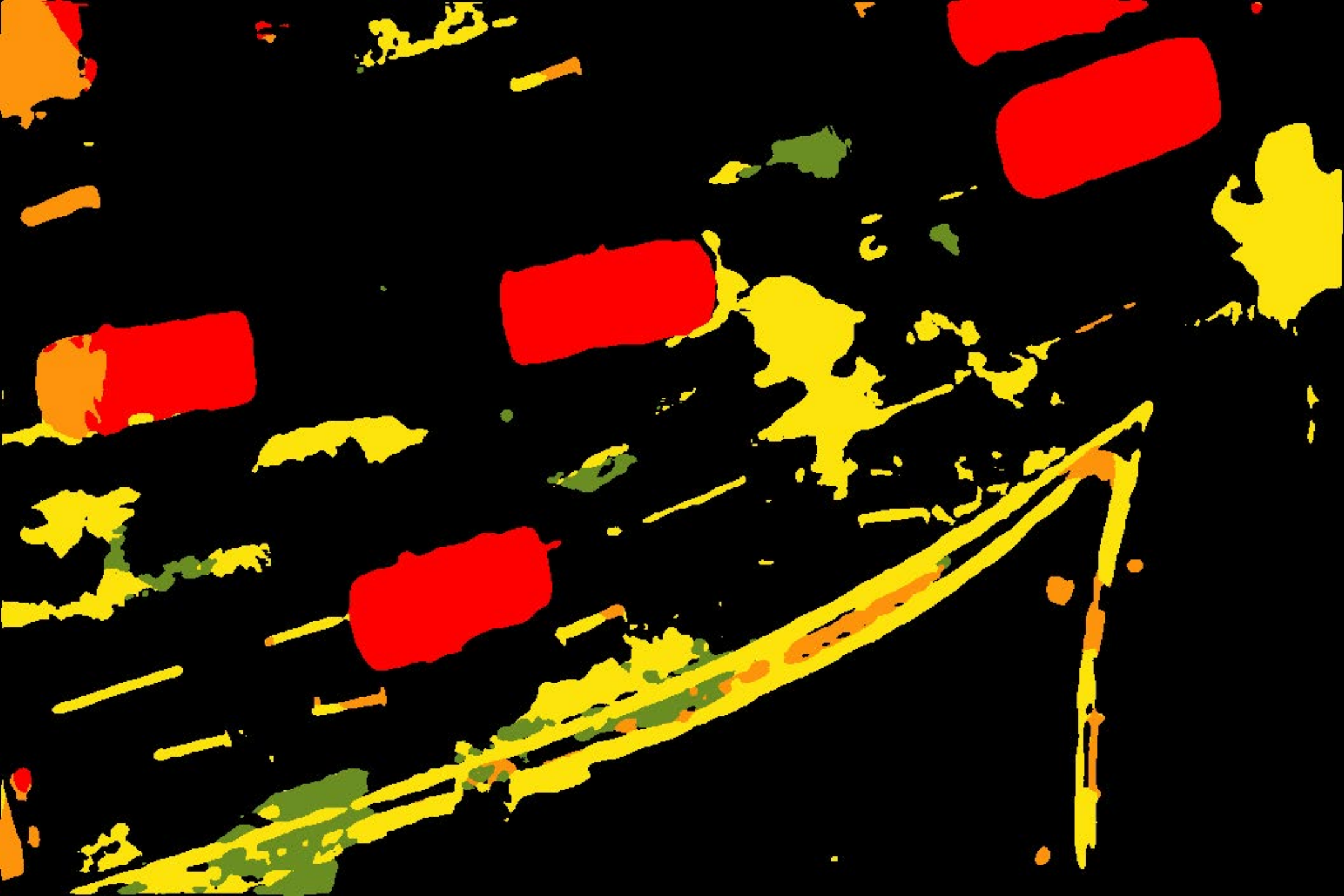}
        \includegraphics[width=.48\linewidth]{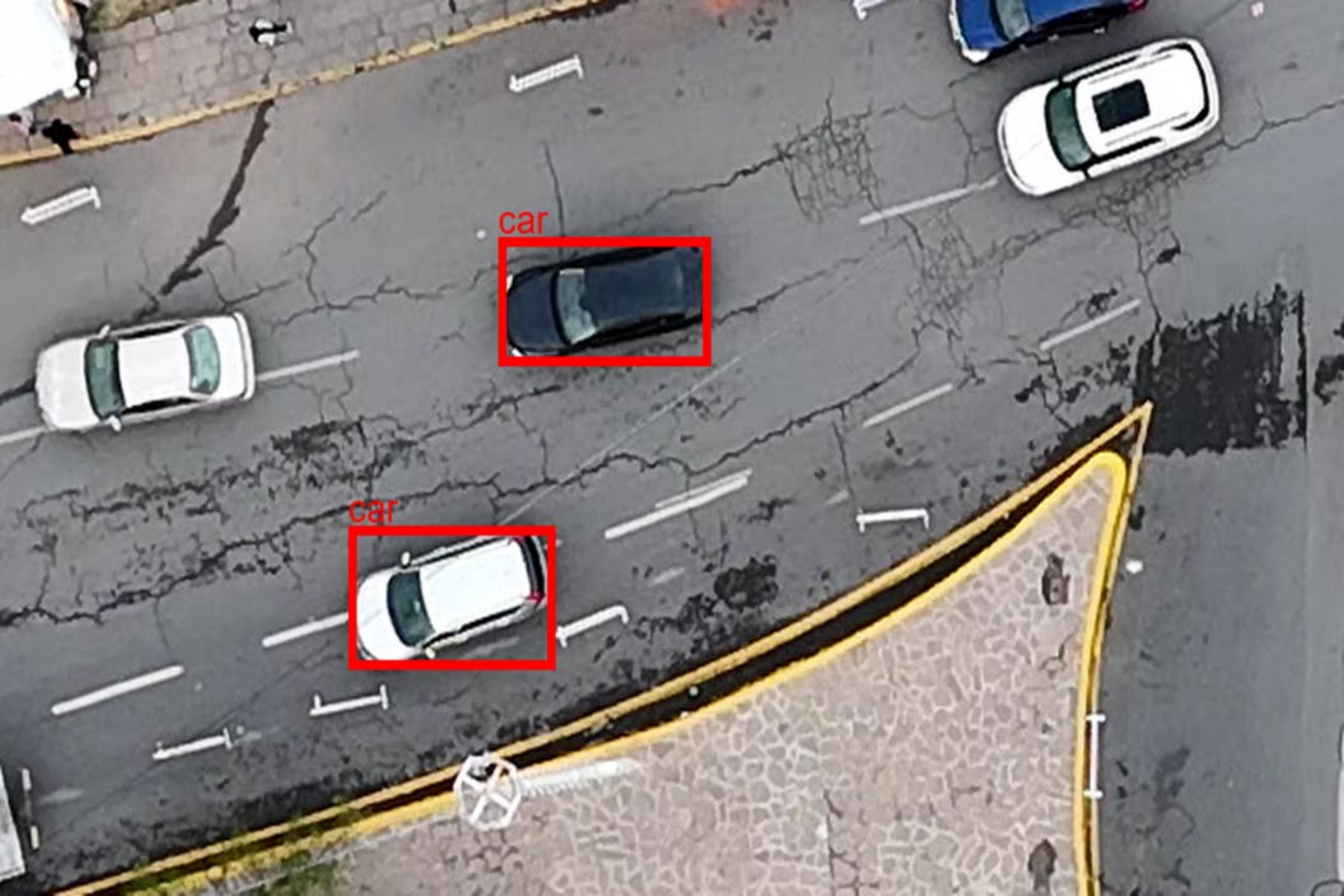}
    \end{subfigure}
    \vspace{0.05 cm}

    \begin{subfigure}{\textwidth}
        \centering
        \includegraphics[width=.48\linewidth]{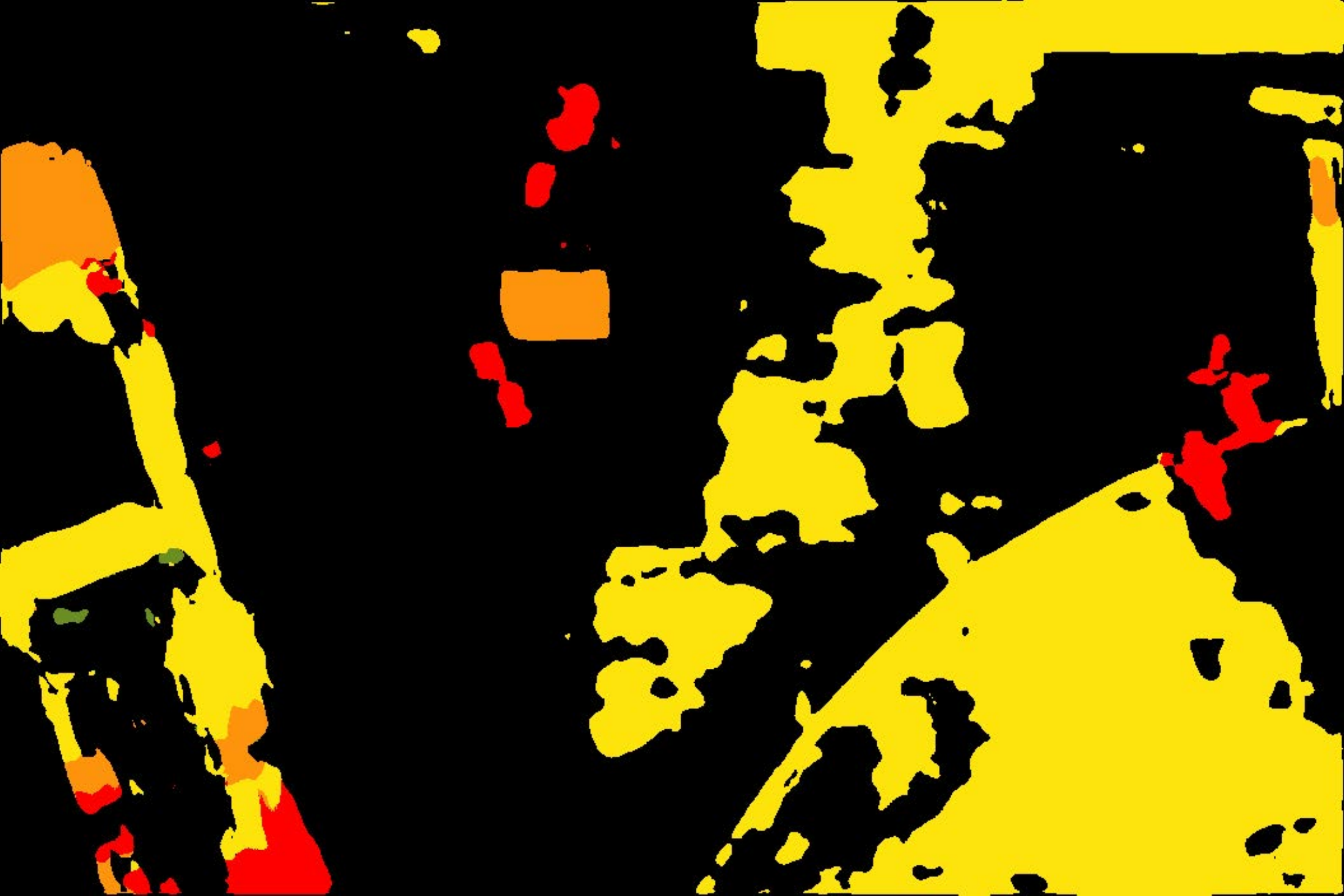}
        \includegraphics[width=.48\linewidth]{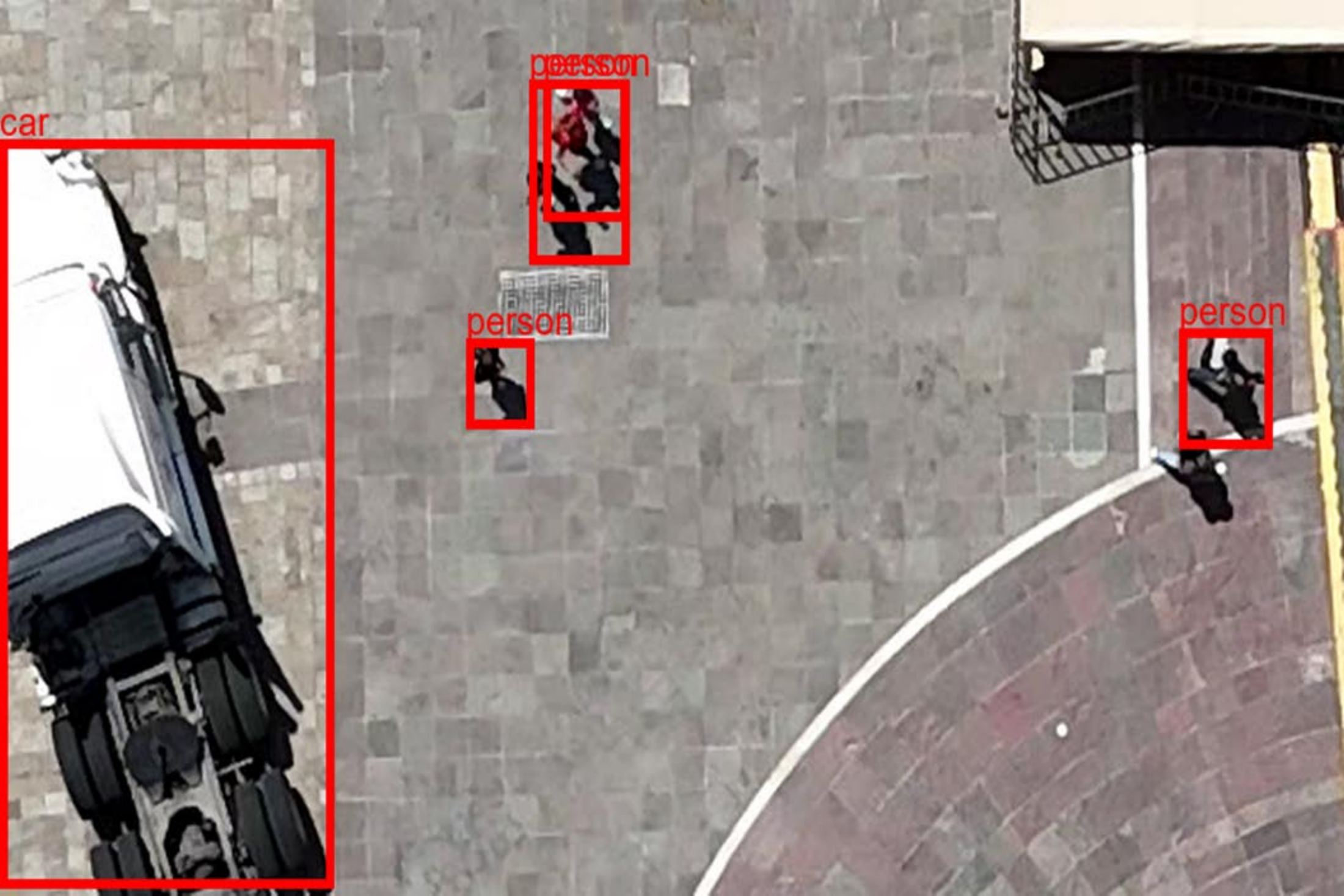}
    \end{subfigure}
    \vspace{0.05 cm}

    \begin{subfigure}{\textwidth}
        \centering
        \includegraphics[width=.48\linewidth]{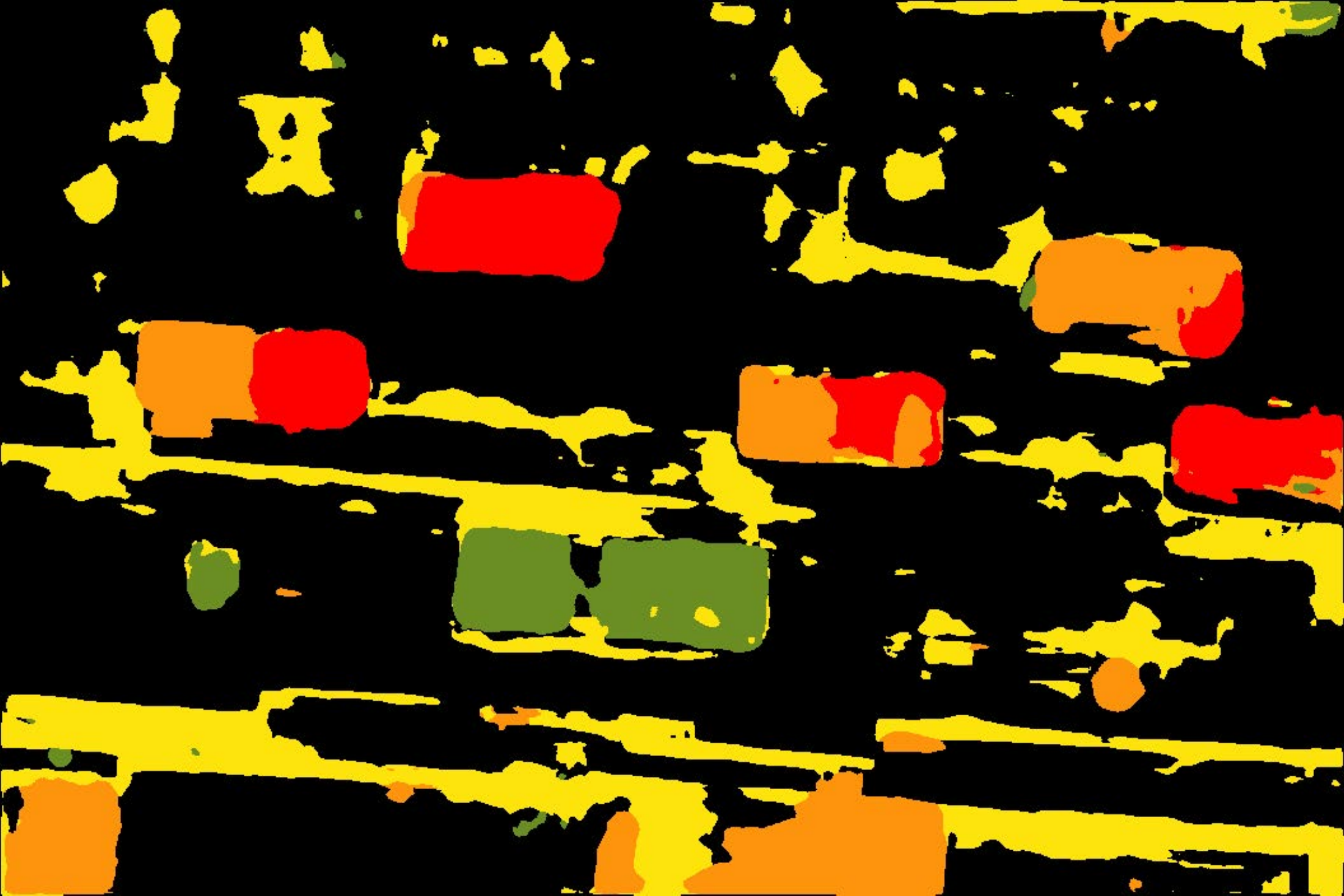}
        \includegraphics[width=.48\linewidth]{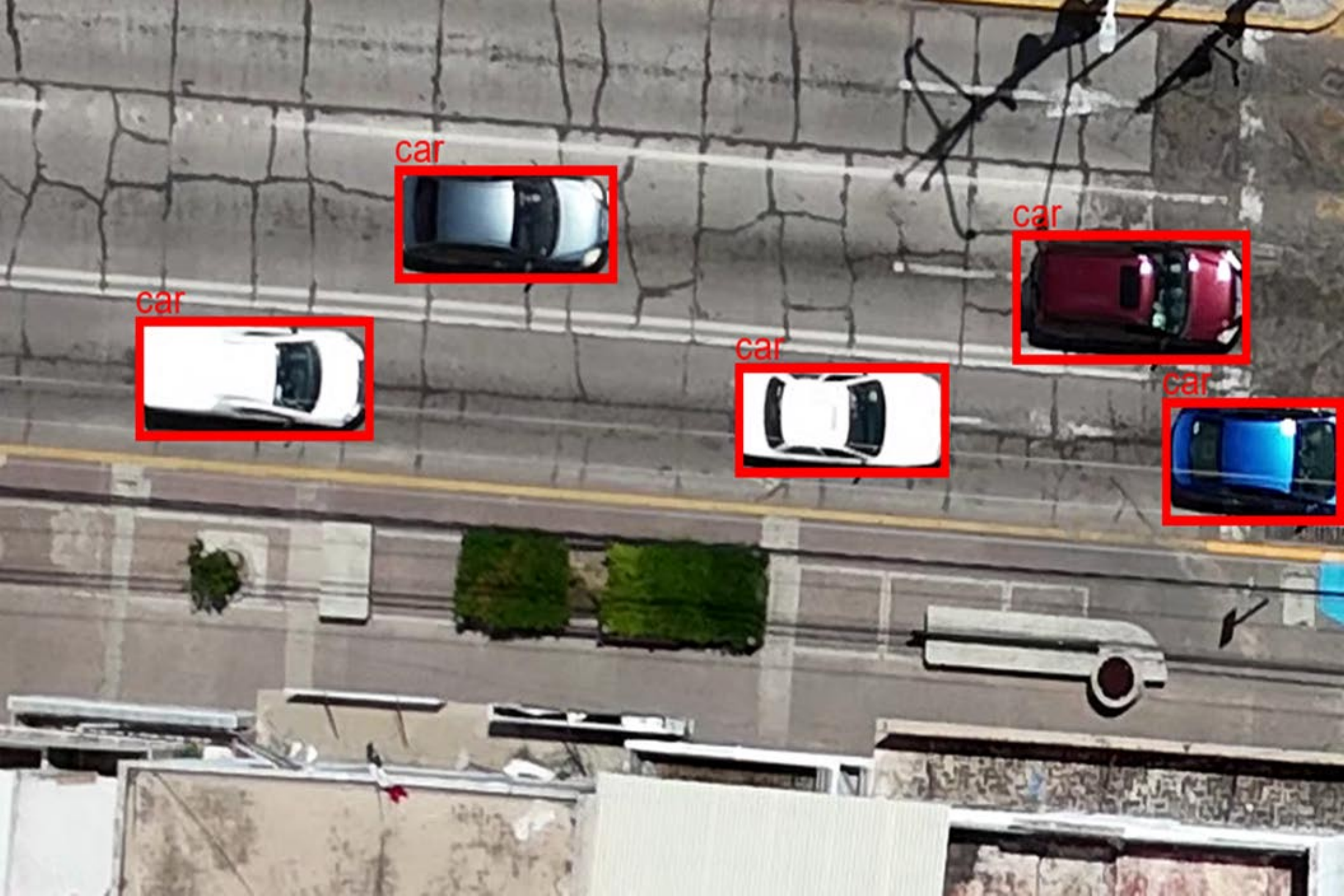}
    \end{subfigure}

    \caption{Performance evaluation of U-Net and YOLO outputs for risk assessment and object detection in UAV landing. U-Net (left) shows low-risk areas in black and high-risk zones, like vehicles and people, in red. YOLO (right) detects risk objects with red bounding boxes.}
    \label{fig:est}
\end{figure}
In autonomous landing algorithms for aerial vehicles, identifying safe zones is crucial to avoid accidents. This requires detecting both risk objects and elements of the environment where the UAV can safely land. For example, it is essential to identify surfaces like pavement, dirt, or grass that are suitable for landing, while also considering the presence of dynamic objects that could cause serious or even fatal collisions.

To showcase the usefulness of the ViVa-SAFELAND framework, the first case study is presented here where a U-Net network was trained using the dataset at \cite{visualization-tools-for-semantic-drone-dataset}, comprising images captured from an aerial perspective at altitudes between 5m and 30m. These high-resolution images (6000$\times$4000 pixels) were divided into a training set of 400 images and a test set of 200 images. Each image includes corresponding segmentation data across 24 classes, enabling the detection of various elements within the visual field. These classes were re-categorized into five risk levels, the lowest level corresponding to areas safe for landing (grass, dirt, etc.), while the highest risk level is designated for critical objects such as people, cars, and bicycles. With the trained U-Net model, the output provides a segmented image indicating the level of risk associated with each detected area. These risk levels are derived from severity levels established by \cite{guerin2021certifying}, which explore safety requirements for emergency landings in compliance with the Specific Operations Risk Assessment (SORA) standard \cite{SORA2019}, from the European legislation. The objective of this model is to pinpoint high-risk zones to avoid during UAV landings, while identifying safe areas such as grass or dirt. These results can serve as input to a decision-making algorithm, or a fully autonomous navigation strategy, helping to select the safest possible landing area based on the calculated risk. It may also serve as an auxiliary tool to train human or autonomous pilots.

The second case study implements a YOLOv8 model to detect moving risky obstacles, such as people, cars, motorcycles, animals, bicycles, etc. This model identifies various high-risk objects within the VAC FoV, providing essential information for real-time environment assessment. Given that these elements pose significant landing risks, the data generated can be integrated into a decision-making model, enabling the selection of safer landing zones and helping to minimize the number of accidents.

Fig. \ref{fig:est} presents a comparison between performance of the two conducted studies. On the left, the output of the U-Net model is shown, where areas in black indicate lower-risk levels, while red areas represent higher-risk zones. On the right side of the same image captured by the VAC, the YOLO model output is displayed, identifying people and vehicles within the FoV by marking them with red bounding boxes to indicate detections. In Sub-figure (a), it can be observed that the YOLO model does not accurately detect all vehicles, and the risk level model erroneously classifies certain areas of the ground as low-risk zones when they should not be. This highlights the significant limitations and errors in current segmentation and detection models, which could result in property damage or even endanger human lives if applied in real autonomous landing scenarios without prior validation. This framework was specifically designed to test and validate models in a controlled environment, as well as to evaluate and select the most suitable strategies and parameters, safeguarding the physical integrity of individuals and avoiding harm to third parties or the UAV itself, until well matured solutions are available and properly tested.

\section{CONCLUSIONS AND FUTURE WORK}
\label{sec:conclusions}
ViVa-SAFELAND, a new open source software tool has been developed for the validation, training, and safe testing of vision based perception and navigation algorithms related to aerial vehicles, with particular interest in real unstructured urban scenarios. This system operates by emulating a UAV and a virtual camera over a real environment previously recorded at a safe altitude, considering its motion dynamics, and ensuring that the people's safety and the vehicle's integrity are not compromised. The framework allows for testing in environments that, for legal and safety reasons, would be inaccessible for experiments with real UAVs.

In two case studies, the framework demonstrated its ability to effectively operate in real scenarios, providing reliable results in the evaluation of moving obstacles detection, and risk assessment algorithms, associated with emergency landings. By simulating the dynamics of a UAV equipped with a camera and enabling the integration of decision-making algorithms, the framework facilitates the faster and safer development of advanced UAV solutions. Moreover, the framework offers a fair comparison baseline for establishing a benchmark to evaluate different navigation strategies over different realistic scenarios. Also, it may be used to help in the automatic generation of labeled datasets for different training tasks. Additionally, it offers a safe platform for human and autonomous pilot training, eliminating the risk of accidents.

Therefore, this work provides an efficient tool for the validation and improvement of vision navigation and autonomous landing algorithms, significantly reducing the risks associated with UAV deployment in densely populated urban environments. It also opens new opportunities for implementing more robust and efficient solutions in civilian UAV applications, with a primary focus on safety and compliance with existing regulations.

As part of future work, the plan is to continue developing and improving the framework to expand its capabilities. One key proposal is the incorporation of advanced algorithms for aerial image segmentation, which will allow for more efficient and automated database generation. This will not only optimize the dataset creation process but also enhance the accuracy and versatility of the framework in validating and training UAV algorithms, particularly for landing and autonomous navigation missions. Furthermore, the development of fully autonomous navigation algorithms using visual feedback is also envisioned.

\section*{acknowledgments}
This work was supported by Office of Naval Research Global ONRG, Award No. N62909-24-1-2001.

\section*{conflict of interest}
The authors declare that there is not conflict of interest of any kind.

\section*{supporting Information}

The ViVa-SAFELAND framework will be released as open source upon acceptance.

A dataset containing the recorded videos in real urban scenarios is provided at: 

\href{https://zenodo.org/records/13942934?preview=1&token=eyJhbGciOiJIUzUxMiJ9.eyJpZCI6ImIzNTllMGYyLTdhMmUtNGMzOS1iMTZmLTY0OTUxYWQ0ZDE1ZSIsImRhdGEiOnt9LCJyYW5kb20iOiJiZWVmMmUzZThiODg4NDQwYzIxZjBkNGM0YWI4YWE4YSJ9.9oYE8d1dDHjXRrJdtzQc43vDaSYg_fAxf9HEcvt83-S1IMN9NUtwUYKVmLnMM3N2JQhWI8qYI99hozAPR5ixbw}{https://zenodo.org/records/13942934?}


\printendnotes

\bibliography{biblio}



\end{document}